\documentclass[a4paper,fleqn]{cas-dc}

\usepackage[authoryear,longnamesfirst]{natbib}

\usepackage{tabularx}
\usepackage{multirow}
\usepackage[dvipsnames]{xcolor}

\usepackage[caption=false]{subfig}
\usepackage{orcidlink}
\def\tsc#1{\csdef{#1}{\textsc{\lowercase{#1}}\xspace}}
\tsc{WGM}
\tsc{QE}
\tsc{EP}
\tsc{PMS}
\tsc{BEC}
\tsc{DE}

\begin{document}
\let\WriteBookmarks\relax
\def\floatpagepagefraction{1}
\def\textpagefraction{.001}

\shorttitle{Tree Boosting Methods for Balanced and Imbalanced Classification}

\shortauthors{Velarde et~al.}

\title [mode = title]{Tree Boosting Methods for Balanced and Imbalanced Classification and their Robustness Over Time in Risk Assessment}                      



%


\author[1]{Gissel Velarde}[orcid=0000-0001-5392-9540]

%
%
\fnmark[1]
%
\ead{Work carried out at Vodafone GmbH. Current Affiliation: International University of Applied Sciences GmbH.  gissel.velarde@iu.org}
%
%
%
\author[1]{Michael Weichert}


\author[1]{Anuj Deshmunkh}
\author[1]{Sanjay Deshmane}
\author[1]{Anindya Sudhir}
\author[1]{Khushboo Sharma}
\author[1]{Vaibhav Joshi}
\affiliation[1]{organization={Vodafone GmbH.},
    Adressaten={Ferdinand Platz 1}, 
    city={Düsseldorf},
    postcode={40549}, 
    country={Germany. michael.weichert@vodafone.com. http://www.vodafone.com}}
\begin{abstract}
Most real-world classification problems deal with imbalanced datasets, posing a challenge for Artificial Intelligence (AI), i.e., machine learning algorithms, because the minority class, which is of extreme interest, often proves difficult to be detected. This paper empirically evaluates tree boosting methods' performance given different dataset sizes and class distributions, from perfectly balanced to highly imbalanced. For tabular data, tree-based methods such as XGBoost,  stand out in several benchmarks due to detection performance and speed. Therefore, XGBoost and Imbalance-XGBoost are evaluated. After introducing the motivation to address risk assessment with machine learning, the paper reviews evaluation metrics for detection systems or binary classifiers. It proposes a method for data preparation followed by tree boosting methods including hyper-parameter optimization. The method is evaluated on private datasets of 1 thousand (K), 10K and 100K samples on distributions with 50, 45, 25, and 5 percent positive samples. As expected, the developed method increases its recognition performance as more data is given for training and the F1 score decreases as the data distribution becomes more imbalanced, but it is still significantly superior to the baseline of precision-recall determined by the ratio of positives divided by positives and negatives. Sampling to balance the training set does not provide consistent improvement and deteriorates detection. In contrast, classifier hyper-parameter optimization improves recognition, but should be applied carefully depending on data volume and distribution. Finally, the developed method is robust to data variation over time up to some point. Retraining can be used when performance starts deteriorating. 
\end{abstract}

 \begin{graphicalabstract}
 \includegraphics[width=17.5cm]{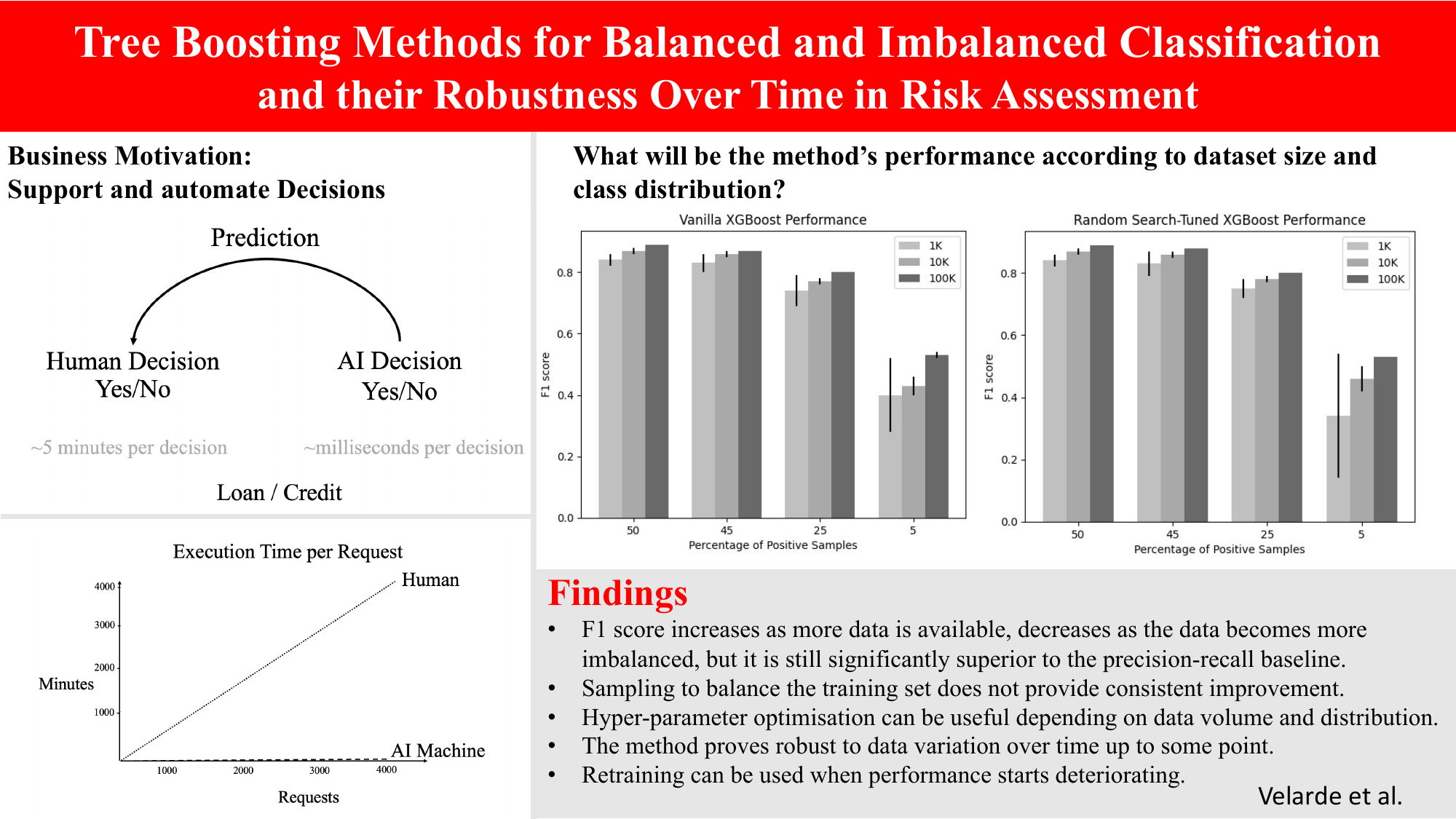}
 \end{graphicalabstract}

\begin{highlights}
\item It describes a Tree Boosting-based method evaluated empirically on balanced and imbalanced datasets of different volumes. 
\item It provides examples to illustrate how different evaluation measures are to be interpreted for detection systems or binary classifiers.
\item On private datasets, it demonstrates empirically that the method increases its detection performance as the data volume increases. Besides, the F1 score decreases as the data distribution becomes more imbalanced, but it is still significantly superior to the baseline of precision-recall or ratio between positives divided by positives and negatives.
\item It shows that sampling to balance the training set to deal with imbalance does not consistently improve recognition. More generally, it worsens detection.
\item It discusses the care that should be taken when considering hyper-parameter optimization depending on data volume and class distribution. 
\item It shows that the developed method is robust to data variation over time, up to some point. Retraining can be used when performance starts deteriorating.
\end{highlights}

\begin{keywords}
Balanced \& Imbalanced Classification \sep XGBoost  \sep Machine Learning  \sep AI  \sep Risk Assessment  \sep Performance Evaluation
\end{keywords}

\maketitle

\section{Introduction} \label{s:introduction}
Classification is a widely applied machine learning task in industrial setups. Outside laboratories, there might be a few cases where class distribution is balanced, since most real-world problems deal with imbalanced datasets. Binary classification systems are evaluated on their ability to correctly identify negative and positive samples. Detecting positive samples is critical, and often challenging when the datasets are imbalanced \citep{lemaitre2017imbalanced, wang2020imbalance, kim2022, hajek2022fraud, li2023imbalanced, yang2021progressive, wang2020imbalance}. 

In risk assessment, positive class samples may represent substantial business losses. At the same time, negative samples are essential, and therefore, flagging a negative sample as positive, is a lost business opportunity. Furthermore, the challenges are the following:

\begin{itemize}
\item positive examples may represent rare cases, they can be anomalous and continuously change their behavior,
\item patterns may even be unseen during training, and 
\item there might be a considerable delay until an abnormal activity is identified. Sometimes, realizing that an activity was anomalous can take months.
\end{itemize}

In 2021, estimations in the telecommunications sector report that loss due to anomalous activity accounts for around USD 40 Billion, representing over two percent of the global revenue of USD 1.8 Trillion \citep{Jacob2021}. These activities include equipment theft, illegitimate commissions, and device reselling, in which global losses were estimated at USD 3.1 Billion, USD 2.2 Billion, and USD 1.7 Billion, respectively \citep{Jacob2021, yang2021progressive}.

In recent years, eXtreme Gradient Boosting (XGBoost) has gained attention since it stands out as a highly competitive approach in machine learning contests for its recognition performance and speed \citep{chen2016xgboost}. In this study, XGBoost is systematically evaluated on small, medium, and large datasets presenting different class distributions. 

The contributions of this paper are the following:
\begin{itemize}
\item It provides examples to illustrate how different evaluation measures can be interpreted for detection systems or binary classifiers.
\item It reviews the principles of Boosting Trees and the advantages of XGBoost as the selected boosting system.
\item It describes a method for a Vanilla XGBoost and a Random Search-Tuned XGBoost.  
\item On private datasets, it demonstrates empirically that the method increases its detection performance as the data volume increases. Besides, the F1 score decreases as the data distribution becomes more imbalanced, but it is still significantly superior to the baseline of precision-recall or ratio between positives divided by positives and negatives.
\item It shows that sampling to balance the training set to deal with imbalance does not consistently improve recognition. More generally, it worsens detection.
\item It discusses the care that should be taken when considering hyper-parameter optimization depending on data volume and class distribution. 
\item It shows that the developed method is robust to data variation over time, up to some point. Retraining can be used when performance starts deteriorating.
\end{itemize}

The following section motivates the automation and decision-making support of risk assessment with AI, i.e., machine learning. Section \ref{s:evaluation} reviews evaluation in binary detection systems or binary classifiers. Section \ref{s:xgboost}, reviews XGBoost. The method is described in section \ref{s:method}. The experiments are presented in section \ref{E1}. Finally, conclusions are drawn in section \ref{s:conclusions}. This paper revisits and extends the content, method, and experiments presented in \citep{velarde2023evaluating}.

\section{On the motivation to automate and support risk assessment with AI}
\begin{figure}
\centering
\includegraphics[width=8cm]{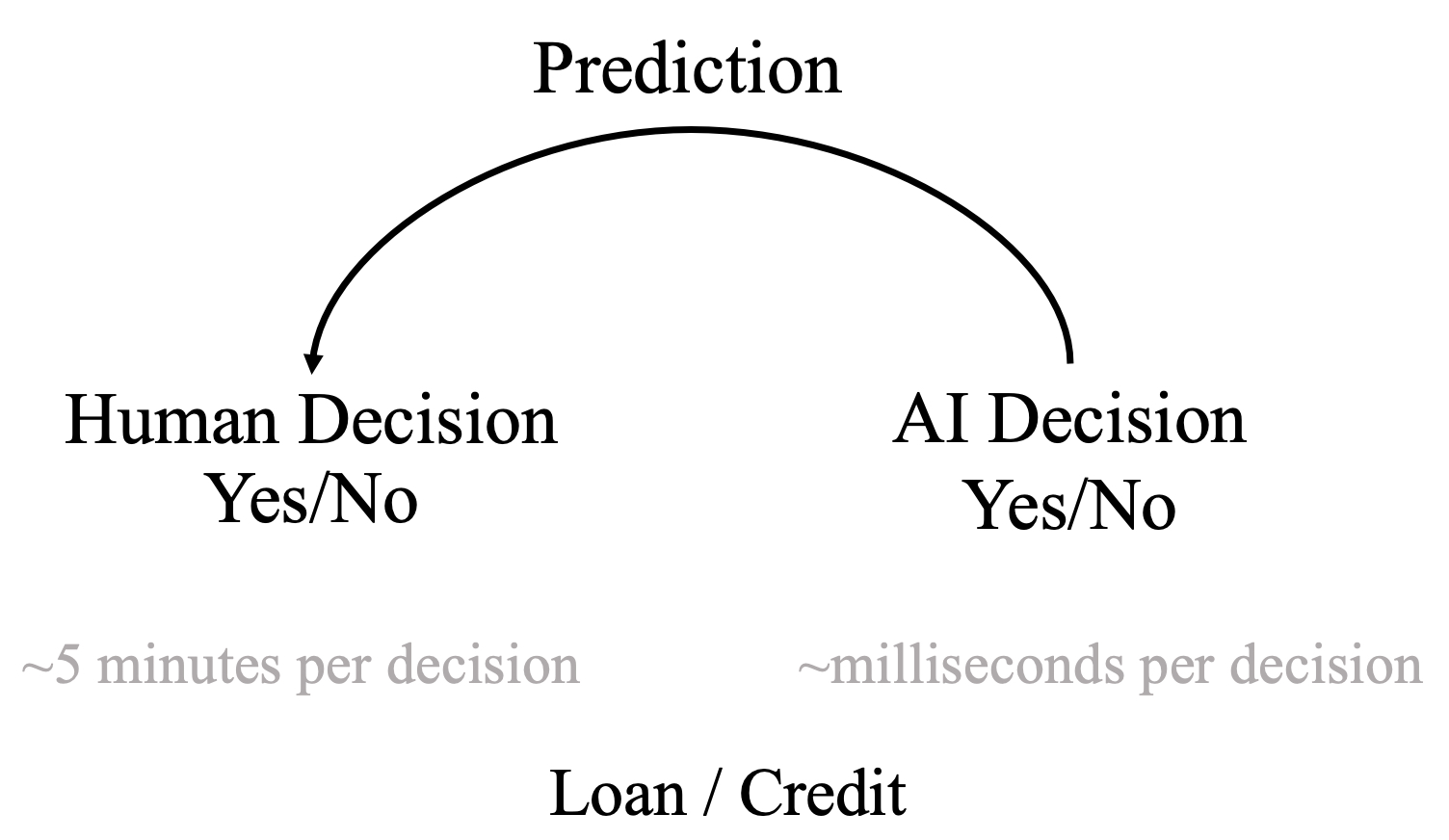} 
\caption{AI can simulate human decisions in a shorter time, helping human inspectors save time and focus on critical cases. From \citep{Scaling_Velarde}.}
\label{fig:loan}
\end{figure}
\begin{figure}
\centering
\includegraphics[width=9cm]{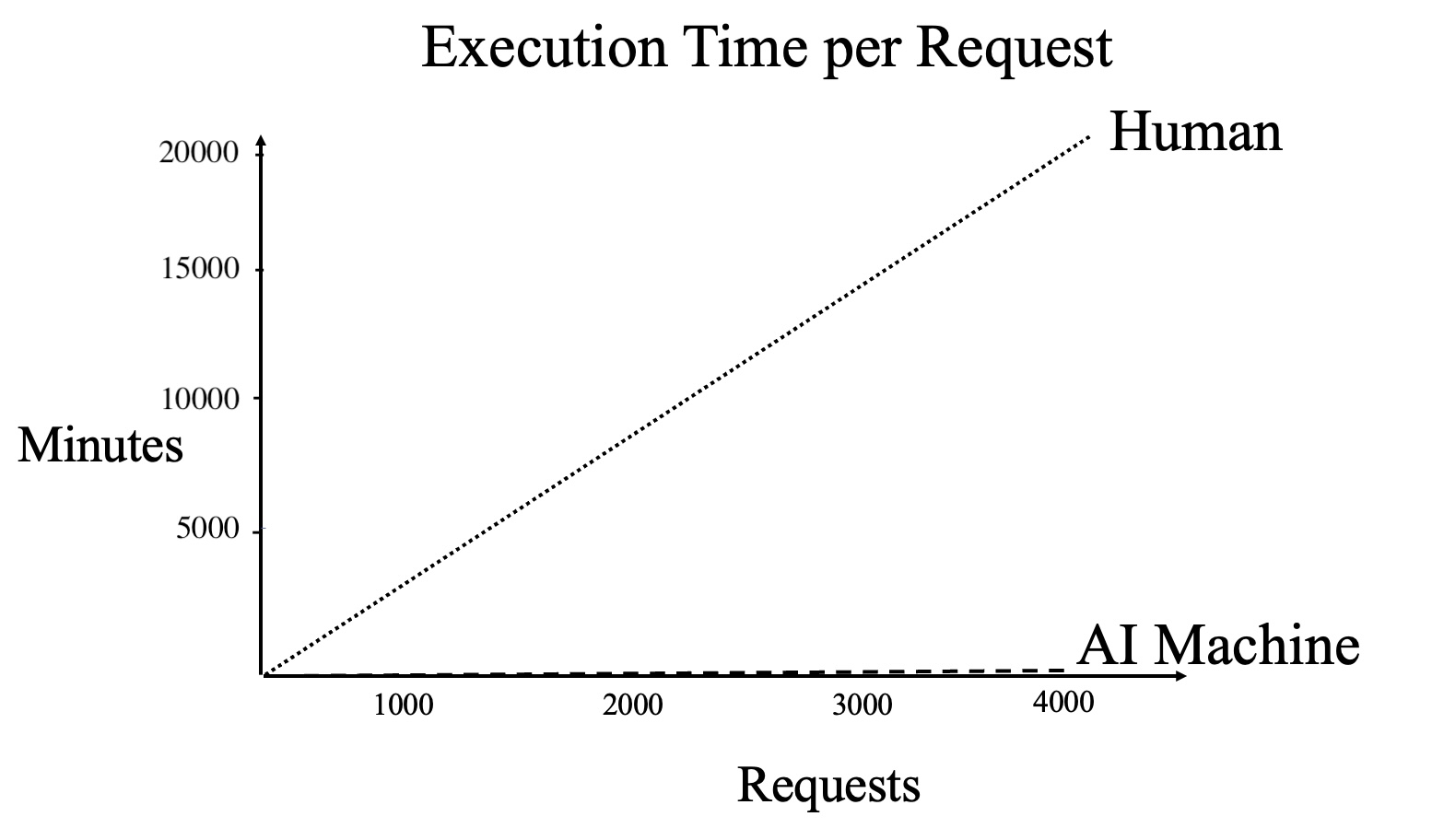} 
\caption{Consider that each decision takes a human expert 5 minutes. Therefore, for 1000 requests, the time to execute will be 5000 minutes, 10000 minutes for 2000 requests, and so on. AI Machines execute each request in milliseconds and, therefore, can help save time, increasing productivity. In addition, AI can easily scale as the number of requests increases. From \citep{Scaling_Velarde}.}
\label{fig:product}
\end{figure}

AI applications have proven to solve a variety of business use cases \citep{WIPO2019}. The question is not whether to apply AI, but where to start to maximize return on investment. In risk assessment, expert inspectors develop, through years and experience, a sense of which customers should be granted a credit or loan. At the same time, they have to carefully screen each new customer and score them to approve or deny a credit or loan for products and services. 

Risk assessment, risk scoring, loan, or credit approval can be supported by AI, where machine learning models, data, and inspectors’ feedback are used to simulate human decisions, see Figure \ref{fig:loan}. Although AI systems are not perfect, they can help human inspectors by automating the process so that they focus on critical cases only. In addition, AI systems allow a store to be available for customers 24/7, therefore, immediately responding to customers' requests and, at the same time, controlling credit approval autonomously. AI systems easily scale. In online stores, there are thousands of daily requests for products. Decision makers can delegate most cases to AI prediction, as it can simulate those learned decisions in a much shorter time, see Figure \ref{fig:product}. 

\section{Evaluation in Detection Systems} \label{s:evaluation}
\begin{figure*}
\centering
\includegraphics[height=9cm]{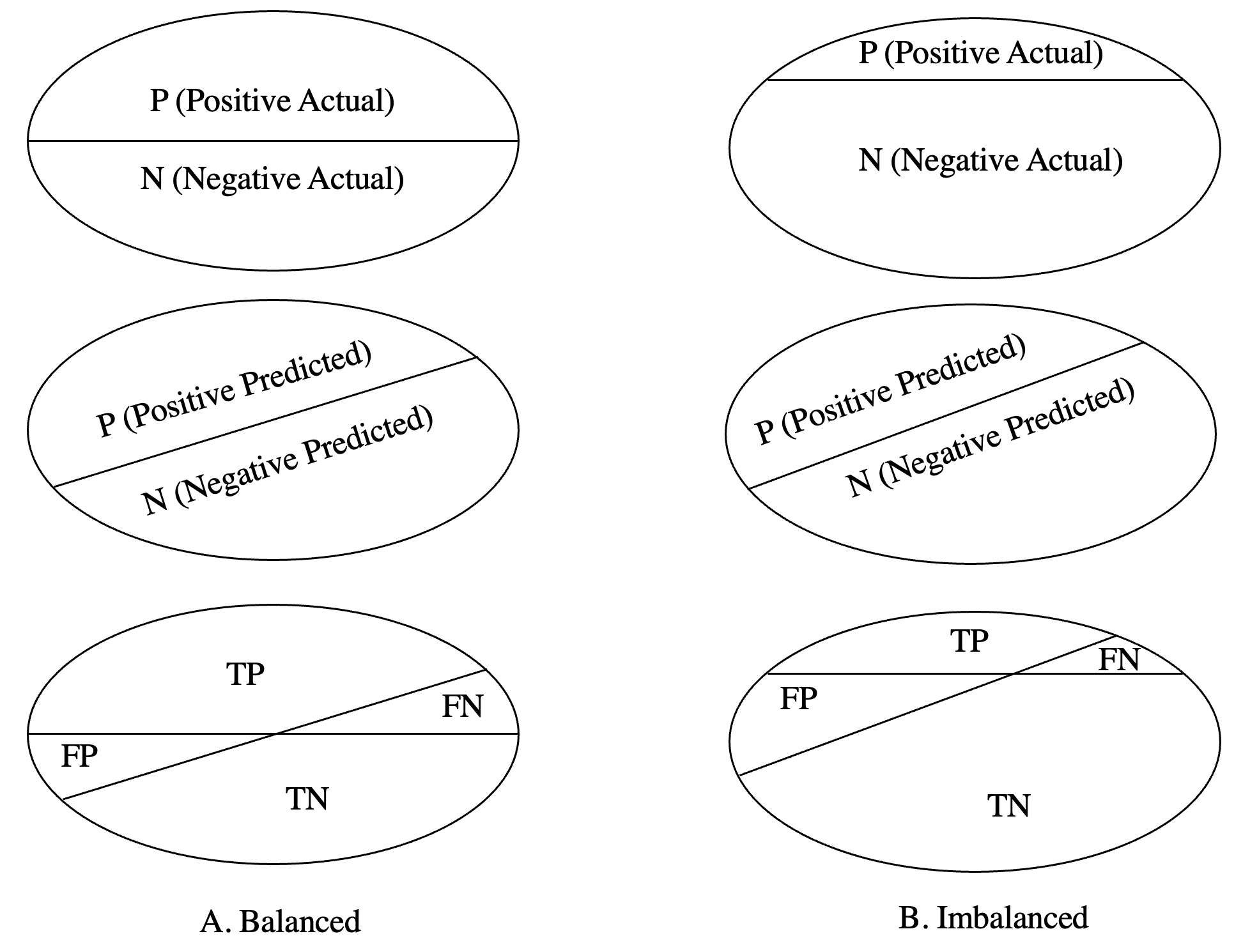}
\caption{Examples of possible distributions for balanced and imbalanced datasets.}
\label{fig:balanced_imbalanced}
\end{figure*}

In detection systems or binary classifiers, we deal with Negative (N) and
Positive (P) samples, where the total number of samples is equal to $N + P$. Detection systems are evaluated considering detection performance, observing the number of True Positive (TP), True Negatives (TN), False Positives (FN), and False Negatives (FN) samples. See Fig. \ref{fig:balanced_imbalanced}. 

Detection systems or classifiers are evaluated considering the following \citep{saito2015precision, chicco2021matthews}: 
\begin{itemize}
\item Confusion Matrix,
\item Area Under Precision-Recall Curve (AUPRC), or (AUC-PR) also known as Precision/Recall Curve (PRC).
\item Precision@n,
\item $F$ scores, depending on the case $F_1$, $F_{0.5}$ or $F_2$,
\item Matthews Correlation Coefficient (MCC),
\item False Positive Rate (FPR), False Negative Rate (FNR),
\item Revenue, or Costs, and
\item Execution time, among other measures.
\end{itemize}

\begin{table*}[]
\begin{center}
\caption{A Confusion Matrix allows us to evaluate if a classifier recognizes or confuses Positive (P) and Negative (N) samples.} \label{t:cm}
\begin{tabular}{llll}
                                                               &                                 & \multicolumn{2}{c}{\textbf{Actual Class}}                                           \\ \cline{2-4} 
\multicolumn{1}{l|}{\textbf{}}                                 & \multicolumn{1}{l|}{\textbf{}}  & \multicolumn{1}{l|}{\textbf{P}}          & \multicolumn{1}{l|}{\textbf{N}}          \\ \cline{2-4} 
\multicolumn{1}{l|}{\multirow{2}{*}{\textbf{Predicted Class}}} & \multicolumn{1}{l|}{\textbf{P}} & \multicolumn{1}{l|}{True Positive (TP)}  & \multicolumn{1}{l|}{False Positive (FP)} \\ \cline{2-4} 
\multicolumn{1}{l|}{}                                          & \multicolumn{1}{l|}{\textbf{N}} & \multicolumn{1}{l|}{False Negative (FN)} & \multicolumn{1}{l|}{True Negative (TN)}  \\ \cline{2-4} 
\end{tabular}
\end{center}
\end{table*}

This paper will focus on $F$ scores, derived from Precision and Recall, as relevant measures to evaluate detection systems. Although the Receiver Operating Characteristic (ROC) curve is still widely used for evaluation, it is only a powerful tool for balanced datasets and is not recommended for imbalanced datasets \citep{saito2015precision}. ROC curves plot $True\ Positive\ Rate\ or\ Recall=TP/(TP+FN)$ versus $False\ Positive\  Rate = FP/(FP+TN)$. When the data is imbalanced, a classifier that commits all possible FP mistakes will look like a good classifier because the TN may substantially outnumber the FP. 

In addition, the developed method optimizes XGBoost using $AUC-PR$ as the evaluation metric. We do not perform thresholding for optimization. Other classifiers, like Logistic Regression, can be tuned to find optimal thresholds on the PR curve, but this is not the case with the developed method. 

The confusion matrix shown in Table \ref{t:cm} allows us to compute  \citep{saito2015precision}:
\begin{equation} \label{eq:1}
Baseline\ PRC = \frac{P}{P+N},
\end{equation}

\begin{equation}
Precision = \frac{TP}{TP+FP},
\end{equation}

\begin{equation}
Recall = \frac{TP}{TP+FN},
\end{equation}

\begin{equation}
F_\beta = (1+\beta^{2})\cdot \frac{Precision \cdot Recall}{\beta^{2}\cdot Precision+Recall},
\end{equation}
where $F_1$ gives same weight to Precision and Recall, $F_{0.5}$ gives more weight to Precision, and $F_{2}$ gives more weight to Recall.

In addition, we can compute:
\begin{equation}
Accuracy = \frac{TP+TN}{TP+FP+TN+FN},
\end{equation}

however, as mentioned before, Accuracy is not recommended when datasets are imbalanced. For instance, see the following examples. 

\subsection{Examples to understand evaluation metrics}
Table \ref{t:1}, presents five examples with $1\,000$ samples each. Example 1 has an equal number of Positive and Negative samples, 500 each, respectively. Examples 2 to 5 have 900 Negative and 100 Positive samples. The Baseline PRC is 0.50 for Example 1 and 0.10 for Examples 2 to 5. 

\subsubsection{Examples 1 and 2.} The classifiers make an equal number of FP and FN mistakes, such that Precision, Recall, $F_1$, $F_{0.5}$, $F_{2}$ are equal to $0.50$. Accuracy is $0.67$ for Example 1 and $0.90$ for Example 2, although the classifier in the second example still makes an equal amount of FP and FN mistakes. 

\subsubsection{Example 3.} It showcases a classifier that flags everything as Negative. In this case, only Recall and Accuracy can be computed, and again, Accuracy gives a misleading score of 0.90. 

\subsubsection{Example 4.} The classifier in this example flags everything as Positive. In this case, Recall is 1. The rest of the measures reflect better the detection ability of such a classifier. As expected, $F_{0.5}$ is worse than $F_1$ and $F_{2}$, because $F_{0.5}$ gives more weight to Recall. 

\subsubsection{Example 5.} It showcases a classifier with high Precision but low Recall. Because this classifier makes no FP mistakes, it is highly precise, but it identifies only five Positive samples out of 100, and therefore its Recall is low. $F$ scores behave as expected. %


The previous examples teach us that while it is important to look at Precision and Recall separately, $F$ scores summarise the performance of classifiers. In addition, Accuracy is deceiving when datasets are imbalanced, and therefore, is not recommended for evaluation.


\begin{table*}[]
\caption{Performance for five detection systems (classifiers) when a dataset is balanced (Example 1) and imbalanced (Examples 2 to 5). Example 1: The classifier makes an equal number of FN and FP mistakes. Example 2: The classifier makes an equal number of FN and FP mistakes. Example 3: The classifier flags everything as Negative; most metrics cannot be computed but Accuracy gives the impression of this classifier being accurate. Example 4: The classifier flags everything as positive and has high Recall but low Precision. Example 5: The classifier has high Precision but low Recall. In red, results that should be observed carefully.} \label{t:1}
\begin{tabular}{llllll}
\textbf{}                       & \textbf{Example 1}          & \textbf{Example 2}          & \textbf{Example 3}          & \textbf{Example 4}          & \textbf{Example 5}          \\ \hline
\textbf{N}                      & 500                         & 900                         & 900                         & 900                         & 900                         \\
\textbf{P}                      & 500                         & 100                         & 100                         & 100                         & 100                         \\ \hline
\textbf{TN}                     & \cellcolor[HTML]{EFEFEF}500 & \cellcolor[HTML]{EFEFEF}850 & \cellcolor[HTML]{EFEFEF}900 & \cellcolor[HTML]{EFEFEF}0   & \cellcolor[HTML]{EFEFEF}900 \\
\textbf{TP}                     & \cellcolor[HTML]{EFEFEF}168 & \cellcolor[HTML]{EFEFEF}50  & \cellcolor[HTML]{EFEFEF}0   & \cellcolor[HTML]{EFEFEF}100 & \cellcolor[HTML]{EFEFEF}5   \\
\textbf{FN}                     & \cellcolor[HTML]{EFEFEF}166 & \cellcolor[HTML]{EFEFEF}50  & \cellcolor[HTML]{EFEFEF}100 & \cellcolor[HTML]{EFEFEF}0   & \cellcolor[HTML]{EFEFEF}95  \\
\textbf{FP}                     & \cellcolor[HTML]{EFEFEF}166 & \cellcolor[HTML]{EFEFEF}50  & \cellcolor[HTML]{EFEFEF}0   & \cellcolor[HTML]{EFEFEF}900 & \cellcolor[HTML]{EFEFEF}0   \\ \hline
\textbf{Precision}              & 0.50                        & 0.50                        &   {\color[HTML]{9A0000} !}                         & {\color[HTML]{9A0000} 0.10}  & {\color[HTML]{9A0000} 1.00}                         \\
\textbf{Recall}                 & 0.50                        & 0.50                        & 0.00                        & {\color[HTML]{9A0000} 1.00}                          & {\color[HTML]{9A0000} 0.50}    \\
\textbf{F1}                     & 0.50                        & 0.50                        &    {\color[HTML]{9A0000} !}  & 0.18                        & 0.10                        \\
\textbf{F0.5}                   & 0.50                        & 0.50                        &    {\color[HTML]{9A0000} !}    & 0.12                        & 0.21                        \\
\textbf{F2}                     & 0.50                        & 0.50                        &     {\color[HTML]{9A0000} !}    & 0.36                        & 0.06                        \\
\textbf{Accuracy}               & 0.67                        & {\color[HTML]{9A0000} 0.90}    & {\color[HTML]{9A0000} 0.90}    & 0.10                        & {\color[HTML]{9A0000} 0.91}                        \\ \hline
\textbf{Baseline PRC}           & 0.50                        & 0.10                        & 0.10                        & 0.10                        & 0.10                       
\end{tabular}
\end{table*}

\section{XGBoost Review} \label{s:xgboost}
\begin{figure*}[]
\caption{Example of a tree ensemble model with two trees. Decision nodes are oval and leaf nodes are rectangular. The numbers inside leaf nodes are scores that contribute to the final prediction. For instance, given an $example$ where $x_1>A$ and $x_3>B$, the final prediction is equal to -1.1 + 1 = -0.1. A convex loss function is used to compare the final prediction with the target to learn the set of functions, minimizing a regularized objective \citep{chen2016xgboost}.} \label{f:exampleTree}
\includegraphics[width=0.9\textwidth]{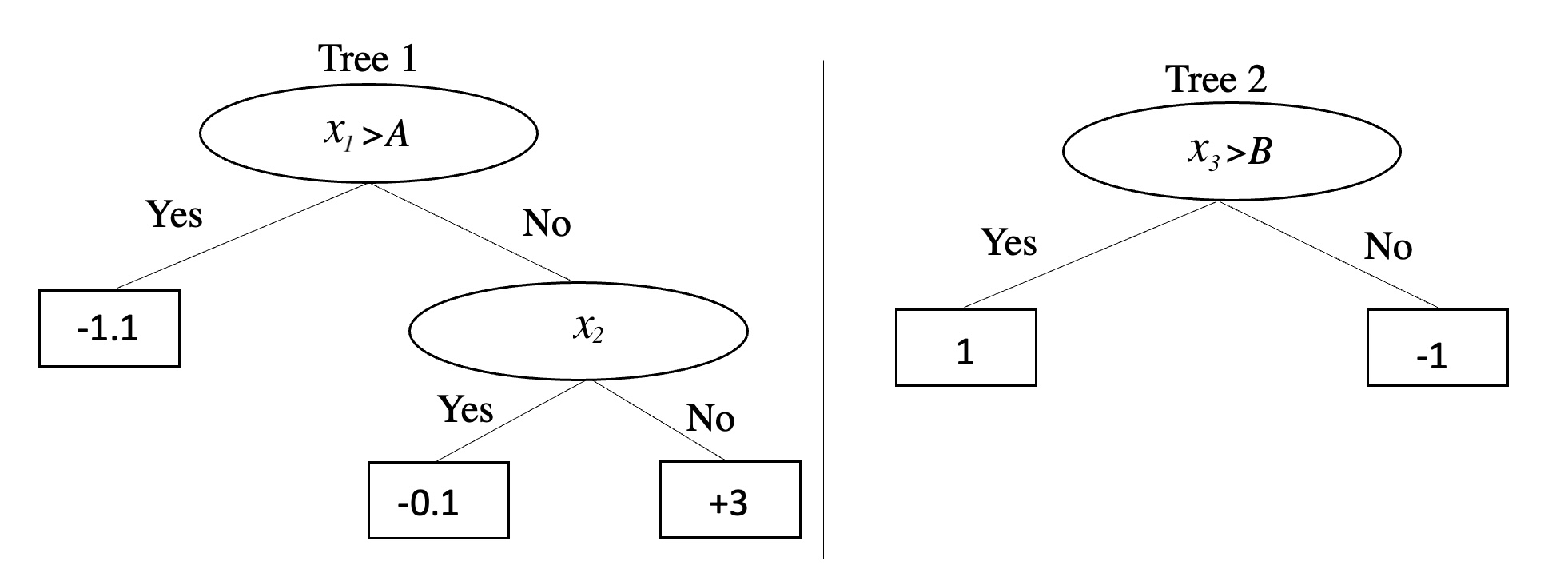}
\end{figure*}

Extreme Gradient Boosting (XGBoost) \citep{chen2016xgboost} is a powerful tree boosting method \citep{friedman2001greedy} that: 
\begin{itemize}
\item First, creates a decision tree,
\item Then, iterates over $M$ number of trees, such that
\begin{itemize}
	\item	It builds a tree likely selecting samples that were misclassified by the previous tree. See Figure \ref{f:exampleTree}.
\end{itemize}
\end{itemize}

\begin{table}[]
\caption{Detection performance for various methods, including supervised, unsupervised, and semi-supervised algorithms on \textit{Paysim} synthetic dataset with more than 6 million samples and nine features where 0.13 percent of samples are positive. Evaluated algorithms include: Extreme Gradient Boosting (XGBoost), Random Forest (RF), Support Vector Machines (SVM), k-Nearest Neighbours (k-NN) and (KNN), Multi-Objective Generative Adversarial Active Learning (MO-GAAL), One-Class SVM (OCSVM), Minimum Covariance Determinant (MCD), Lightweight On-Line detector of Anomalies (LODA), Autoencoder (AE), Variational Autoencoder (VAE), Cluster-Based Local Outlier Factor (CBLOF), Angle-Based Outlier Detection (ABOD),  Histogram-Based Outlier Detection (HBOS), eXtreme Gradient Boosting Outlier Detection (XGBOD). Results in descending F1 score. The best F1 score and execution time per category are in bold. Notice that labels are required to train XGBOD. Data from \protect\citep{hajek2022fraud}.} \label{t:2}
\begin{tabular}{lll}
\textbf{Supervised learning}   & \textbf{F1}     & \textbf{Execution time (s)} \\ \hline
XGBoost                        & \textbf{0.841}  & \textbf{207.0}                \\
RF                             & 0.8394          & $1\,196.2$                     \\
SVM                            & 0.4655          & $12\,082.9$                    \\
k-NN                           & 0.1588          & $4\,581.4$                     \\
\textbf{Unsupervised learning} &                 &                             \\ \hline
MO-GAAL                        & \textbf{0.6059} & $13\,184.4 $                   \\
OCSVM                          & 0.273           & $802.9  $                     \\
KNN                            & 0.1260           & $1\,948.5$                     \\
MCD                            & 0.1084          & $127.4 $                      \\
LODA                           & 0.1060           & $14.8$                        \\
AE                           & 0.0869          & $931.1$                       \\
VAE                          & 0.0869          & $2\,922.9  $                    \\
CBLOF                          & 0.0822          & $41.3$                        \\
ABOD                           & 0.0680           & $2\,646.5$                     \\
Isolation Forest               & 0.0189          & $189.9$                       \\
HBOS                           & 0.0077          & $\textbf{4.1}$                \\
\textbf{Semi-supervised}       &                 &                             \\ \hline
XGBOD                          & \textbf{0.8737} & $\textbf{4\,256.3}$            \\
\end{tabular}
\end{table}

 On tabular data, XGBoost has been reported to win several machine learning competitions \citep{chen2016xgboost}. Related work shows that XGBoost outperforms several machine learning algorithms in mobile payment transactions \citep{hajek2022fraud}. The authors used a synthetic dataset called \textit{Paysim} containing more than 6 million samples with nine attributes or features, where only 0.13 percent of samples are positive. The study evaluated several supervised, unsupervised, and semi-supervised learning algorithms, reporting that Random Forest (RF) obtains the second best $F_1$ for supervised learning algorithms, being XGBoost much faster than RF. Besides, XGBOD, which is presented as a semi-supervised learning method performs best, but to train XGBOD, labels are required just like in a supervised setting, and XGBOD is very slow compared to XGBoost. Multi-Objective Generative Adversarial Active Learning (MO-GAAL) returns the fourth-best score, and it is the slowest of all algorithms. See Table \ref{t:2}. 

Among boosting systems \citep{chen2016xgboost}, XGBoost turns out to be the fastest in comparison to H20 and Spark MLLib and possesses several characteristics, like sparsity awareness and parallel computing, that made it our choice for experimentation. For example, the Exact Greedy Algorithm deals with finding the bests splits and enumerates all of them overall features, being computationally expensive. Thus, XGBoost implements approximate local (per split) and global (per tree) solutions to the problem of finding the best splits \citep{chen2016xgboost}.

  


\section{The method} \label{s:method}
The method is based on a pipeline that prepares the data before XGBoost classification, either with a Vanilla XGBoost with default parameters, a Random Search (RS)-Tuned XGBoost over a set of parameters, or a Grid Search-Tuned XGBoost on the scale parameter. 

\subsection{Data preparation}
Data was prepared as follows:
\begin{itemize}
\item Numerical data was scaled between 0 and 1. 
\item Categorical data was encoded with an ordinal encoder, such that values unseen during training received a reserved value of -1.
\end{itemize}

Although, in theory, XGBoost does not need numerical scaling, we found empirically that scaling improves recognition when the dataset is large. We don't have a theoretical explanation of this effect. 

\subsection{Vanilla XGBoost} \label{s:vanilla}
A Vanilla XGboost, with most default parameters \citep{xgboost_doc}, was tested with the following setup:

\begin{itemize}
\item Binary logistic objective function,
\item Handling of missing values by replacing them with the value of 1,
\item Evaluation metric: AUC-PR, 
\item Maximum depth of a tree equal to 6,
\item Learning rate equal to 0.3,
\item Subsample ratio of training instances before growing trees equal to 1,
\item Subsampling of columns by tree equal to 1, and 
\item Number of trees equal to 100.
\end{itemize}

\subsection{Random Search (RS)-Tuned XGBoost} \label{sec:RS-XGboost}
In addition, Random Search (RS)-Tuned XGBoost models were obtained in cross-validation over the following space:
 
\begin{itemize}
\item Maximum depth of a tree in values equal to 3, 6, 12, and 20,
\item Learning rate in values equal to 0.02, 0.1, and 0.2,
\item Subsample ratio of training instances before growing trees equal to 0.4, 0.8, and 1,
\item Subsampling of columns by a tree in values equal to 0.4, 0.6, and 1, and 
\item Number of trees equal to 100, 1000, and 5000.
\end{itemize}

This set of parameters was tested over random search, given that a winning Kaggle entry on a related task reports optimizing over those parameters \citep{xgboost_tuningDeotte}.

\subsection{Random Search (RS)-Tuned Scale XGBoost} \label{sec:RS-XGboost2} 
In addition, the model was optimized using random search for the scale parameter known as $scale\_pos\_weight$ over the following values: \\
(1, int(75/25), int(95/5), 100, 1000, int(95*100/5)).



\section{Experiments} \label{E1}
The experiments aim to study the method's performance in relation to dataset size and class distribution and its robustness over time. Imbalance classification is a known problem in machine learning~\citep{lemaitre2017imbalanced}. Initial experiments were performed to select possible techniques to deal with imbalance, including sampling techniques and imbalance optimization. Then, the method's robustness over time was studied. Next, the datasets are described.

\subsection{Datasets} \label{dataE1}
\begin{figure*}[]
\centering
\caption{Illustration of the datasets' size.} \label{f:e1}
\includegraphics[width=0.9\textwidth]{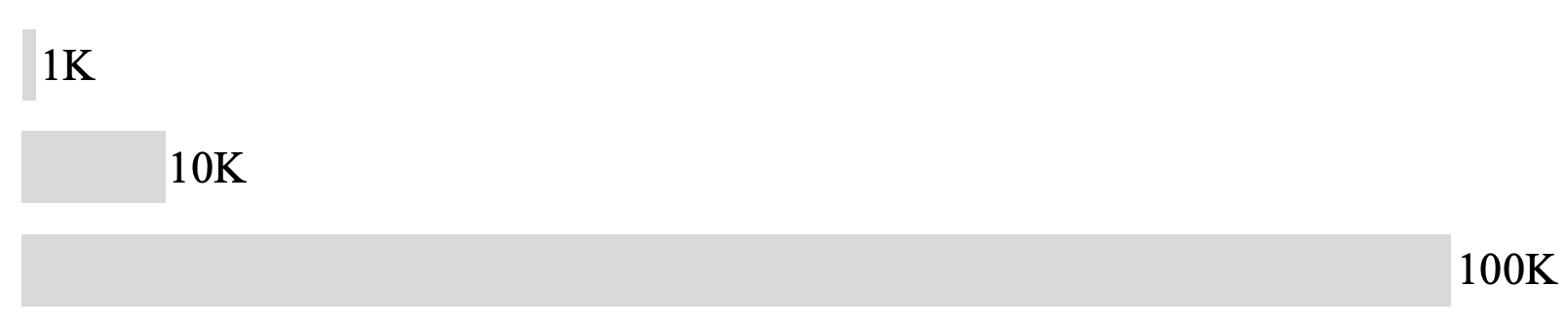}
\end{figure*}

\begin{figure*}[]
\centering
\caption{Illustration of datasets in four distributions: 50\%-50\%, 55\%-45\%, 75\%-25\%, and 95\%-5\%. These were created in sizes of 100K, 10K and 1K samples, as illustrated in Figure \ref{f:e1}. } \label{f:e1b}
\includegraphics[width=0.8\textwidth]{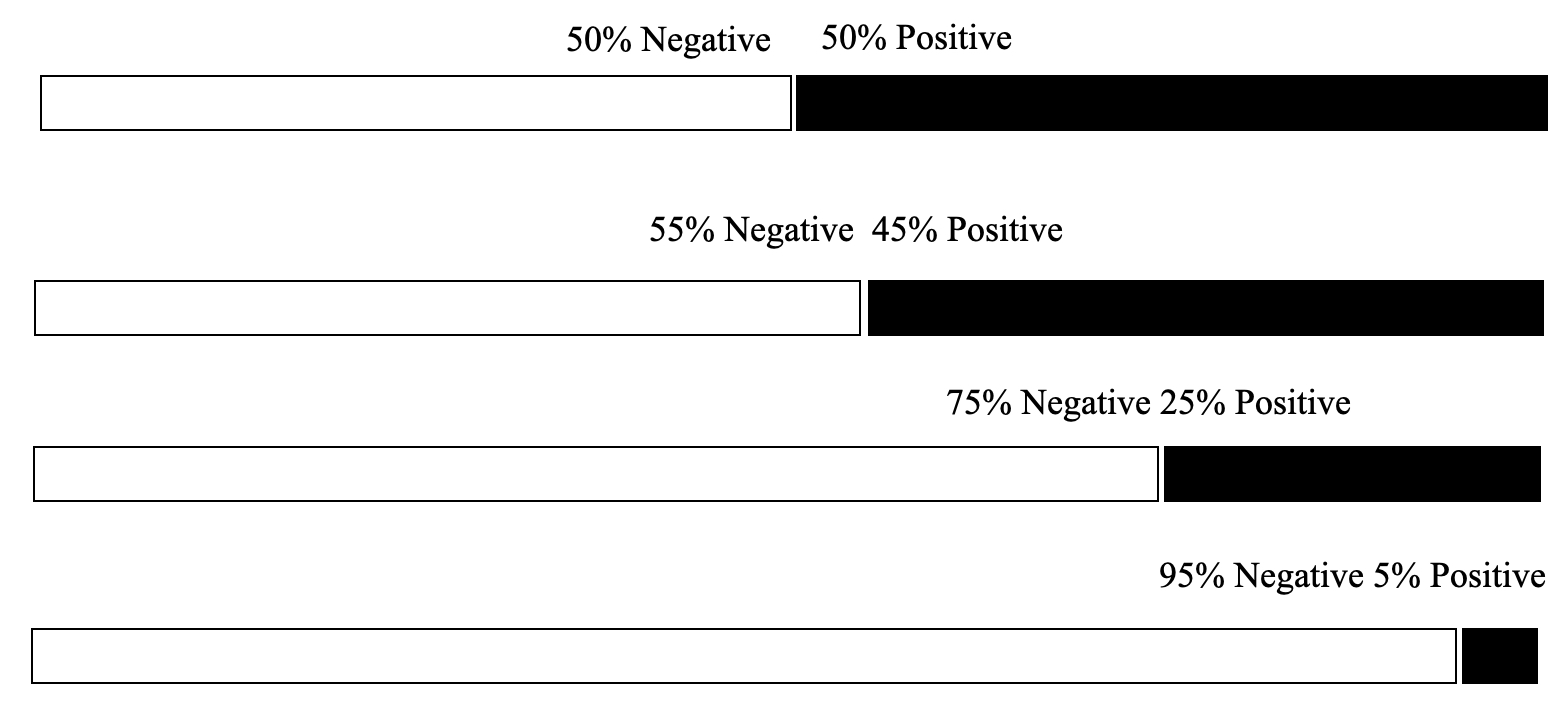}
\end{figure*}

From a large and private dataset with 114 features and 300 thousand (K) samples, datasets of 100K, 10K and 1K samples were created, see Figure \ref{f:e1}, such that there were four distributions from balanced to highly imbalanced in terms of Negative\%-Positive\%: 50\%-50\%, 55\%-45\%, 75\%-25\%, to 95\%-5\%, see Figure \ref{f:e1b}.

\subsection{Sampling techniques to balance the training set} \label{exp.2}
\begin{figure*}[]
\begin{center}
\caption{Sampling techniques aim at balancing or obtaining a balanced number of Positive (P) and Negative (N) samples either by under-sampling the majority class, over-sampling the minority class, or combining under and oversampling.} \label{t:samplingtechniques}
\includegraphics[width=0.65\textwidth]{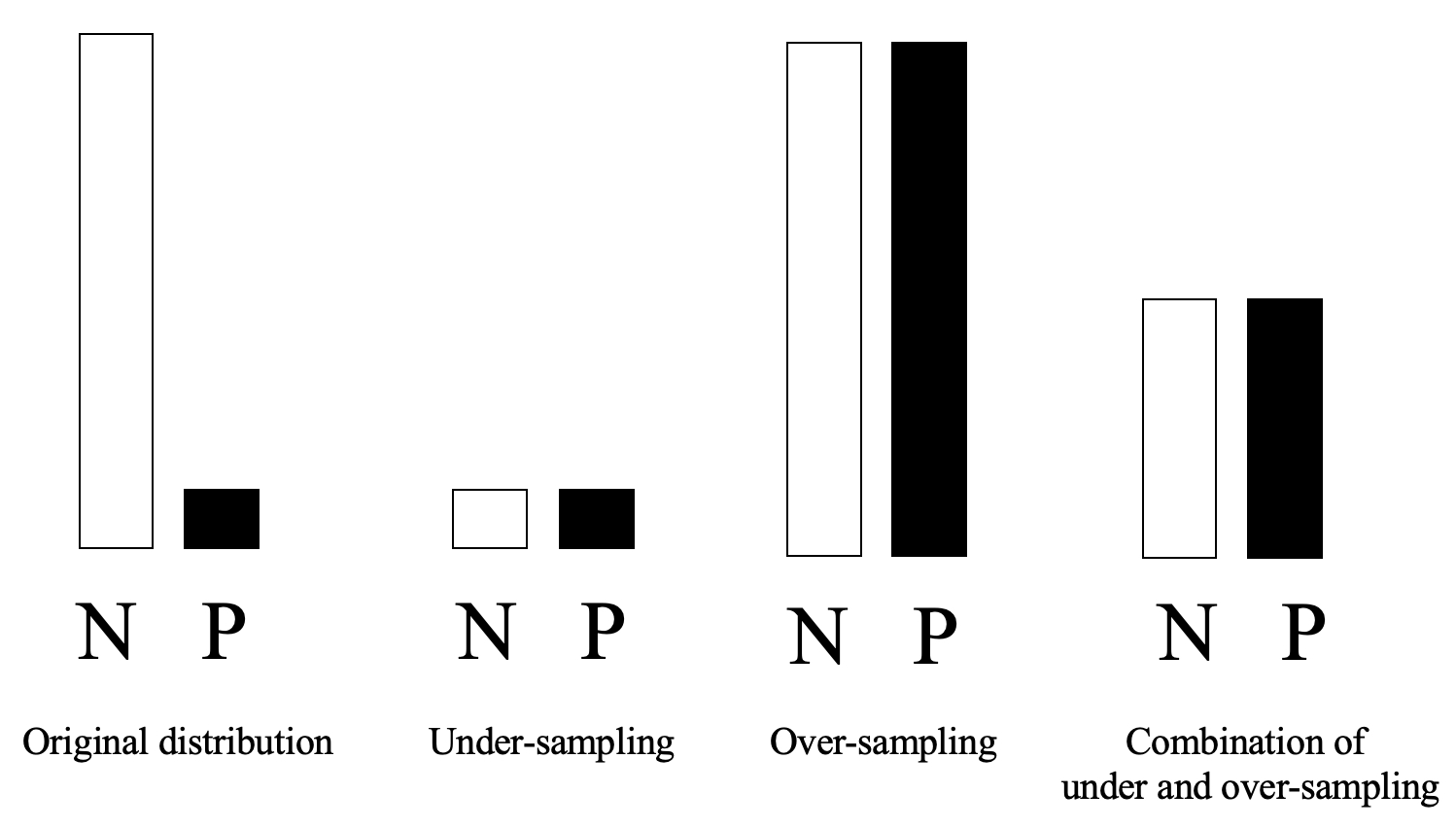}
\end{center}
\end{figure*}

Sampling techniques for imbalance handling include: under-sampling, over-sampling, and a combination of under-sampling and over-sampling \citep{lemaitre2017imbalanced}. See figure \ref{t:samplingtechniques} for illustration. For example, a straightforward approach for under-sampling is Random Under Sampling (RUS). Although Synthetic Minority Over-Sampling Technique (SMOTE) can be used for over-sampling and in combination with under-sampling \citep{lemaitre2017imbalanced}, it was not used for the experiments reported here. From a large dataset, sampling to balance was performed in a combination of under-sampling and over-sampling with real, not synthetic data, always maintaining the size of the datasets. Next, the experiment in section \ref{exp.2} is designed to study the effect of sampling to balance the training set.

\begin{figure*}[]
\begin{center}
\caption{Sampling to balance the training set. (a) 50\%-50\%, (b) 55\%-45\%, (c) 75\%-25\%, and (d) 95\%-5\%. In each case, only the training set was sampled to balance classes. The distribution in the test set is untouched. Consider that the dataset to create the training sets is large enough for sampling to balance.} \label{f:E2}
\includegraphics[width=1\textwidth]{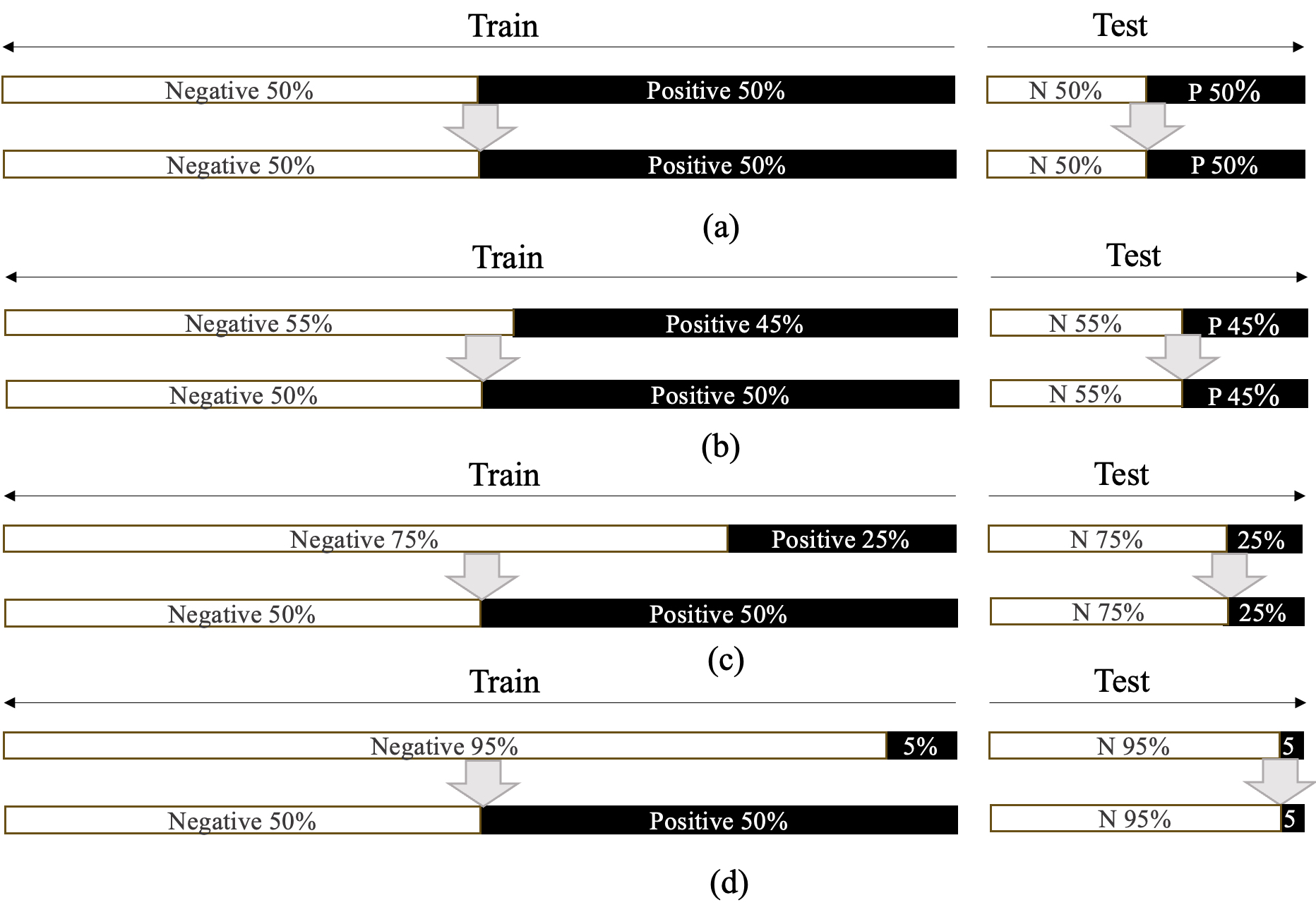}
\end{center}
\end{figure*}

Figure \ref{f:E2} illustrates how the data was sampled, such that each training set had an equal number of positive and negative samples, while the test set reflected four original distributions: 50\%-50\%, 55\%-45\%, 75\%-25\%, and 95\%-5\%. A time point was selected to split the data on training and test sets. Then, sampling followed to obtain the desired distribution for each set and partition. This procedure was applied to a dataset of 10K samples. The experiment was conducted in 80\%-20\%, train-test partition.

Table \ref{t:t4} shows the effect of sampling to balance the training set. Sampling to balance the training set did not produce consistent improvement on F1 scores nor before Vanilla or RS-Tuned XGBoost classification. We observed that besides  F1 deterioration, the relation between precision and recall changed, such that recall improved but precision worsened when sampling to balance the training set. 

Related studies also report the infectiveness of sampling techniques to improve recognition \citep{hajek2022fraud,kim2022}. Random under-sampling (RUS) was reported to deteriorate classifier performance \citep{hajek2022fraud}. Given that random under-sampling reduces the size of the training set, it could be assumed that recognition worsens because the classifier was presented with a smaller training set. However, in the experiments presented here, sampling to balance the training set, did not reduce its size. Nor synthetic positive samples were used to balance the training set, but real samples. Similarly, a study over 31 datasets evaluated the effect of different sampling techniques and found that these were either ineffective or even harmful \citep{kim2022}. Therefore, the idea of continuing testing with several methods for data preparation to deal with imbalance was discarded. 

\begin{table*}[]
\caption{F1 scores. Effect of sampling to balance the training set to 50\%-50\% (Negative - Positive). Dataset size 10K, partition 80-20. The first column shows the percentage of positive samples in the test set. Compare columns 2 and 3; the best results are in bold. The difference between columns 2 and 3 is that the training set in column 5 was sampled to balance before classification with a Vanilla XGBoost. Compare columns 4 and 5; the best results are in bold. The difference between columns 4 and 5 is that the training set in column 5 was sampled to balance before classification with an RS-Tuned XGBoost. In general, sampling to balance the training set deteriorated F1 scores. } \label{t:t4}
\begin{tabular}{|p{2.2cm}|p{2.2cm}|p{2.2cm}|p{2.2cm}|p{2.2cm}|} 
\hline
\textbf{Positive \%} & \textbf{Vanilla XGB} & \textbf{Sampling \& Vanilla XGB} & \textbf{RS-Tuned  XGB} & \textbf{Sampling \& RS-Tuned XGB} \\ \hline
\textbf{50}          & \textbf{0.8731}      & 0.8648               & \textbf{0.8804}        & 0.8728                            \\ \hline
\textbf{45}          & \textbf{0.8721}      & 0.8473               & \textbf{0.8717}        & 0.8535                            \\ \hline
\textbf{25}          & 0.7710               &  \textbf{0.7940}    & \textbf{0.7893}        & 0.7833                            \\ \hline
\textbf{5}           & \textbf{0.4276}      & 0.3973               & \textbf{0.3942}        & 0.3824                            \\ \hline
\end{tabular}
\end{table*}

\subsection{Imabalance-XGBoost optimization for imbalance} \label{sec:imbsgboost}
Imbalance-XGBoost \citep{wang2020imbalance} is an XGBoost's extension with weighted and focal losses that aims at dealing with imbalance learning. Initial experiments were performed on a Diabetes dataset containing 768 samples and 8 attributes, where 34 percent of the samples are positive \citep{diabetesdataset}. 

Imabalance-XGBoost provides two mechanisms to deal with imbalance: weighted-XGBoost and focal-XGBoost. Table \ref{t:imb} shows the comparison in 5-fold cross-validation with a Vanilla XGBoost. Optimization was performed over the following focal gamma ($\gamma$) values: 1, 2, and 3. In addition, optimization was performed over the following weighted-XGBoost alpha ($\alpha$) with values: 1, 2, 3, and 4. For this dataset, Imabalance-XGBoost  focal $\gamma$ deteriorates recognition, while weighted $\alpha$ outperforms Vanilla XGBoost. 

Despite improved recognition when tuning Imbalance-XGBoost over weighted ($\alpha$), we did not continue working with this package as we encountered some difficulties, mostly incompatibility with Scikit-learn \citep{pedregosa2011scikit}. For example, when using Imbalance-XGBoost, the default scoring function seems to be accuracy, and it is not possible to set F1 as a scoring function on Scikit-learn, such that we had to implement our own cross-validation strategy. Besides, Imabalance-XGBoost \citep{wang2020imbalance} does not support missing values as  XGBoost \citep{chen2016xgboost} does. Although we could have used inpainting techniques to deal with missing values, we noticed that Imbalance-XGBoost is incompatible with Scikit-learn pipelines \citep{pedregosa2011scikit}, as it returned errors. Therefore, we stopped experimentation with Imbalance-XGBoost and moved to explore XGBoost optimization for imbalance.  
\begin{table*}[]
\caption{Comparison between Vanilla XGBoost \citep{chen2016xgboost} and Imbalance-XGBoost (Imb-XGBoost) \citep{wang2020imbalance}, F1 scores, mean and standard deviation in 5-fold cross-validation on Diabetes Dataset \citep{diabetesdataset}, containing 768 samples and 8 attributes, where 34 percent of positive samples.} \label{t:imb}
\begin{tabular}{|l|l|l|}
\hline
\textbf{Vanilla XGBoost} & \textbf{Imb-XGBoost, focal $\gamma=1$}    & \textbf{Imb-XGBoost, weighted $\alpha=4$} \\ \hline
0.59 (0.03)                     &       0.55 (0.04)               & \textbf{ 0.65 (0.04)} \\ \hline
      
\end{tabular}
\end{table*}

\subsection{XGBoost optimization for imbalance} \label{sec:finetune}
\begin{figure}[]
\begin{center}
\caption{F1 scores  from Table \ref{tab:cv}. } \label{f:compa0}
\includegraphics[width=0.5\textwidth]{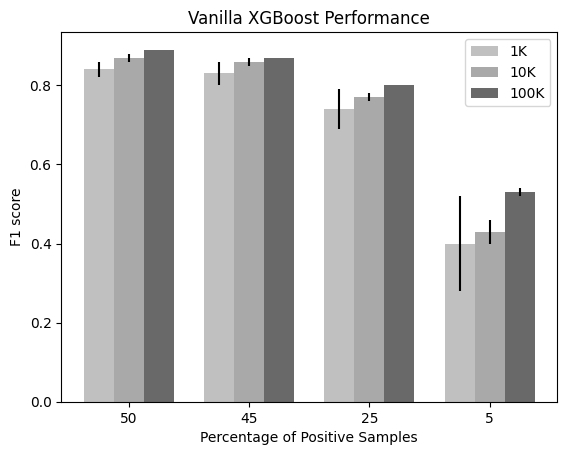}
\end{center}
\end{figure}

\begin{figure}[]
\begin{center}
\caption{F1 scores from Table \ref{tab:cv}. } \label{f:compa}
\includegraphics[width=0.5\textwidth]{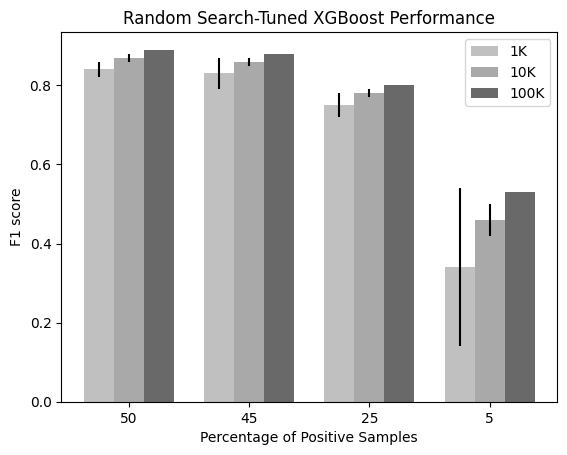}
\end{center}
\end{figure}

\begin{figure}[]
\begin{center}
\caption{F1 scores from Table \ref{tab:cv2}. } \label{f:compa2}
\includegraphics[width=0.5\textwidth]{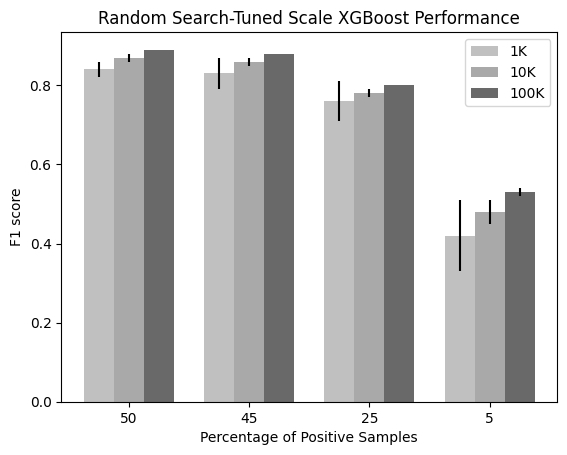}
\end{center}
\end{figure}

\begin{figure}[]
\begin{center}
\caption{F1 scores from Table \ref{tab:cv2}. } \label{f:compa3}
\includegraphics[width=0.5\textwidth]{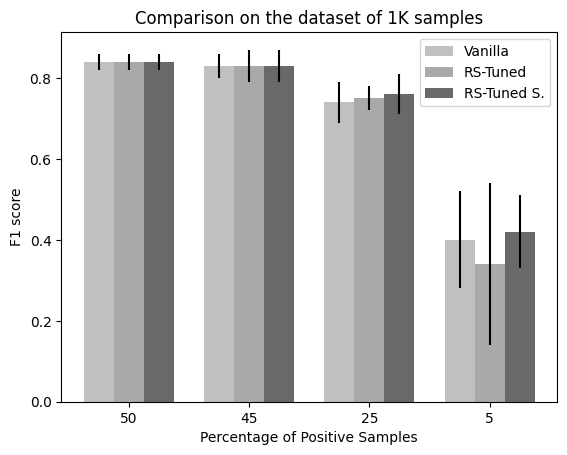}
\end{center}
\end{figure}

\begin{figure}[]
\begin{center}
\caption{F1 scores from Table  \ref{tab:cv2}. } \label{f:compa4}
\includegraphics[width=0.5\textwidth]{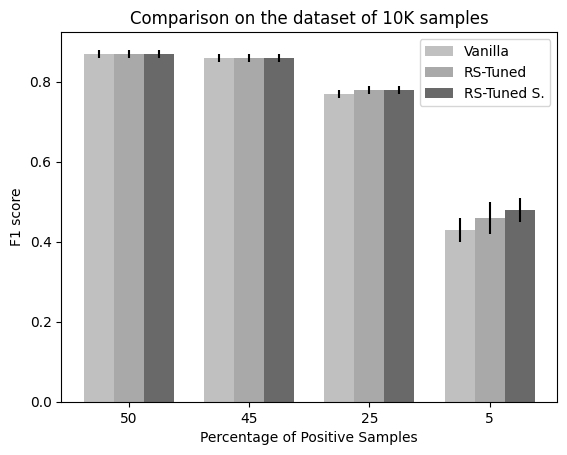}
\end{center}
\end{figure}


This section aims at understanding the method's performance for balanced and imbalanced datasets of different sizes and the impact of hyper-parameter optimization with random search in cross-validation (CV), given that random search proofs more efficient than grid search \citep{bergstra2012random}. The datasets used are those described in section \ref{dataE1}. 

The three approaches listed in section \ref{s:method} were evaluated for datasets of 1K, 10K, and 100K samples in four distributions including 50, 45, 25 and 5 percent of positive samples, that is: Vanilla XGBoost (section \ref{s:vanilla}), RS-Tuned XGBoost optimized over the parameters listed in section \ref{sec:RS-XGboost} and RS-Tuned Scale XGBoost (RS-Tuned S. XGBoost) optimized over the scale parameter, see section \ref{sec:RS-XGboost2}. Figures \ref{f:compa0}, \ref{f:compa}, and \ref{f:compa2} present the results for Vanilla XGBoost, RS-Tuned XGBoost, and RS-Tuned S. XGBoost, respectively.  Figures \ref{f:compa3} and \ref{f:compa4} compare the approaches for the datasets of 1K and 10K, respectively. Alternatively, see Table \ref{tab:cv2}, which summarises the results.

As expected, the method improves $F_1$ score as the dataset increases in size and decreases $F_1$ score as the data distribution becomes more imbalanced. Still, it is significantly better than the baseline of precision-recall  or $Baseline\ PRC$, even for the lowest $F_1$ score of 0.34 (0.2) (two-tailed unpaired t-test at 95\% confidence interval, $t_4$ = 3.2423, p-value = 0.0118). Moreover, the larger and more balanced the dataset, the more stable the classification results. Surprisingly, when the dataset is perfectly balanced, even with the smallest dataset of 1K, it is possible to achieve a similar performance to that of the largest dataset of 100K. As the data distribution becomes more imbalanced, hyper-parameter optimization should be performed carefully, more so if the datasets are small, as it could damage performance. Hyper-parameter optimization over the parameters listed in section \ref{sec:RS-XGboost} is not recommended for the smallest dataset of 1K samples, when the dataset  distribution has 5 percent positive samples, since the average F1 score worsens and the standard deviation gets larger. In contrast, optimising over the \textit{scale\_pos\_weight}, described in section \ref{sec:RS-XGboost2}, proves satisfactory, see Figures \ref{f:compa3} and \ref{f:compa4}. Unexpectedly, for this particular dataset, there is no difference between a Vanilla XGBoost and the optimized pipeline when the dataset's size is 100K samples.

\begin{table*}
\centering
\caption{Summary of results for Vanilla XGBoost (section \ref{s:vanilla}), RS-Tuned XGBoost (section \ref{sec:RS-XGboost}), and RS-Tuned S. XGBoost (section \ref{sec:RS-XGboost2}). Five-fold cross-validation (CV) for datasets of 1K and 10K samples, and 2-fold cross-validation for datasets of 100K samples in different distributions (Distr.) of 50, 45, 25, and 5 percent positive samples. Second column: Baseline PRC. The small and medium datasets in the distribution of 25 and 5 percent positive samples are the most benefited by optimizing the scale parameter.}
\label{tab:cv2}
\begin{tabular}{|l|c|l|c|l|} 
\hline
\textbf{Dataset size}             &                   & 1K        & 10K      & 100K             \\ 
\hline
\textbf{Cross-validation} &                   & CV=5        & CV=5        & CV=2               \\ 
\hline
\textbf{Distribution}   & \textbf{Baseline PRC} & \multicolumn{3}{c|}{\textbf{Vanilla XGBoost}}  \\ 
\hline
\textbf{50-50}                    & 0.5               & 0.84 (0.02) & 0.87 (0.01) & 0.89 (0.00)        \\ 
\hline
\textbf{55-45}                    & 0.45              & 0.83 (0.03) & 0.86 (0.01) & 0.87 (0.00)        \\ 
\hline
\textbf{75-25}                    & 0.25              & 0.74 (0.05) & 0.77 (0.01) & 0.80 (0.00)        \\ 
\hline
\textbf{95-05}                    & 0.05              & 0.40 (0.12) & 0.43 (0.03) & 0.53 (0.01)        \\
\hline

                                 & \textbf{Baseline PRC} & \multicolumn{3}{c|}{\textbf{RS-Tuned XGBoost}}  \\ 
\hline
\textbf{50-50}                    & 0.5               & 0.84 (0.02)               & 0.87 (0.01) & 0.89 (0.00)        \\ 
\hline
\textbf{55-45}                    & 0.45              & 0.83 (0.04)               & 0.86 (0.01) & 0.88 (0.00)     \\ 
\hline
\textbf{75-25}                    & 0.25              & 0.75 (0.03)               & 0.78 (0.01) & 0.80 (0.00)       \\ 
\hline
\textbf{95-05}                    & 0.05              & 0.34 (0.20)               & 0.46 (0.04) & 0.53 (0.00)      \\
\hline

                                 & \textbf{Baseline PRC} & \multicolumn{3}{c|}{\textbf{RS-Tuned S. XGBoost}}  \\ 
\hline

\textbf{50-50}                    & 0.5               &   0.84 (0.02)                  & 0.87 (0.01) & 0.89 (0.00)    \\ 
\hline
\textbf{55-45}                    & 0.45              &      0.83 (0.04)                  & 0.86 (0.01) & 0.88 (0.00)    \\ 
\hline
\textbf{75-25}                    & 0.25              &    0.76 (0.05)                  & 0.78 (0.01) & 0.80 (0.00)     \\ 
\hline
\textbf{95-05}                    & 0.05              &   0.42 (0.09)                  & 0.48 (0.03) & 0.53 (0.01)    \\
\hline

\end{tabular}
\end{table*}

\label{tab:cv}

\subsection{Robustness to data variation over time} \label{sec:robust}
\begin{figure*}[htp]
\centering
\subfloat[]{%
  \includegraphics[width=12.5cm]{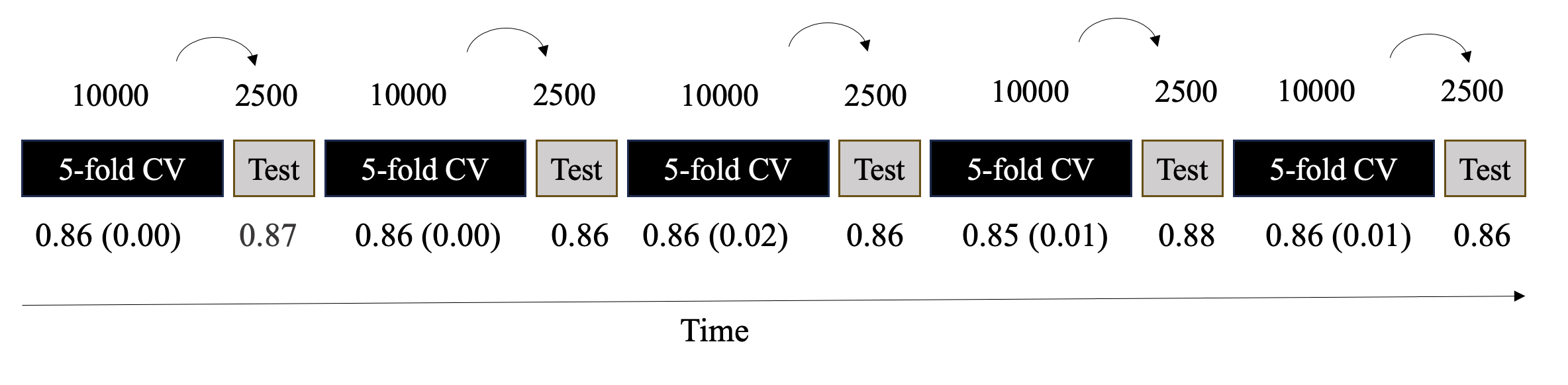} %
}

\subfloat[]{%
  \includegraphics[width=12.5cm]{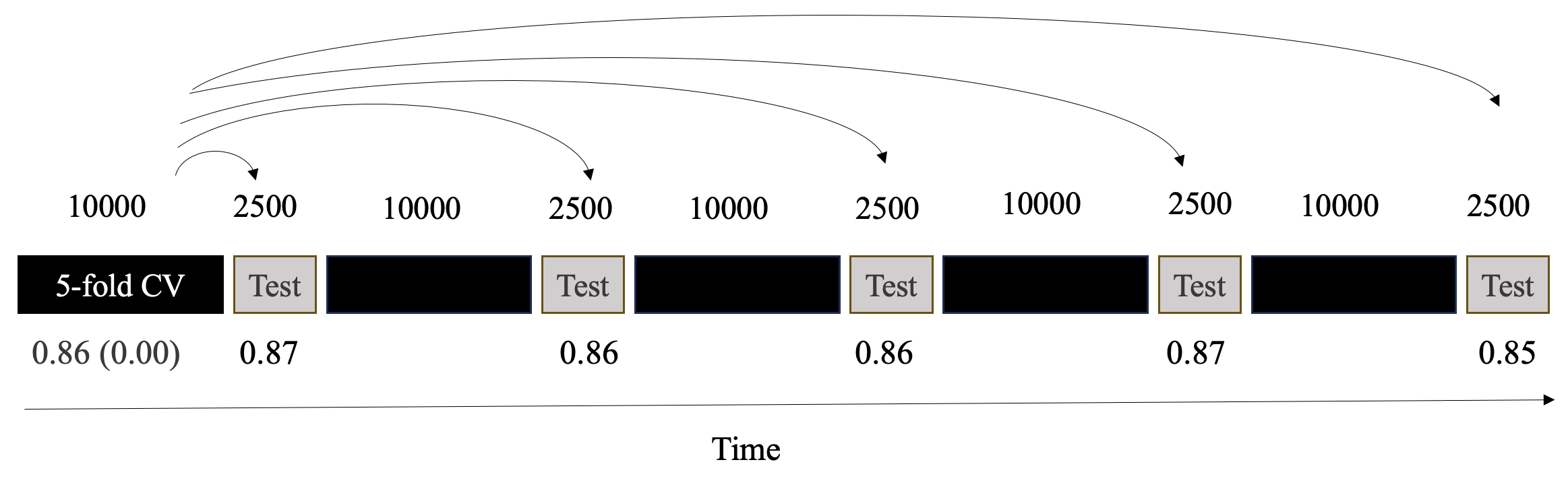}%
}

\caption{(a) Training and testing like in a moving window over time. Training sections contain 10000 samples, test sections 2500 samples. Black sections are used to find the optimal model by hyper-parameter optimization in 5-fold Cross-Validation. Once the model is found, the 10000 samples are used to train the optimal model, and its immediate gray section is the test set.  The average $F1= 0.87\ (0.01)$ on the 5 test sets in gray, when the training set is the entire previous set of 10000 samples (black). (b) Training only once and testing over time. The first section is trained with hyper-parameter tuning in 5-fold Cross-Validation, and then all 10000 samples in the first chunk are used to train the best model found. In this setup, only the first chunk of data is used for training and used to predict all future test sections. The average $F1=0.87\ (0.01)$ in (b) is equal to that in (a).}\label{fig:time}

\end{figure*}
Expert inspectors describe that from time to time, new patterns emerge, such that new strategies are developed and recognized as new anomalous patterns. To test the robustness of the model to data variation over time, the following experiment was performed, see Figure \ref{fig:time} for illustration. The distribution of the dataset is  55\%-45\%.  

First, chunks of $10\,000$ samples were used for training and the next $2\,500$ samples for testing. 
In Figure \ref{fig:time} (a), the black sections correspond to model training and gray to testing, such that each model is evaluated on its immediate test section. The average performance over the 5 test sections is $F1= 0.87\ (0.01)$. Once an optimised model was found in cross-validation, the 10000 samples were used for training the best model to predict the immediate test section.

In Figure \ref{fig:time} (b), the setup is different as in Figure \ref{fig:time} (a). In this case, only the first black section is used to train and select the best model which is used to predict all gray test sections. The average $F1=0.87\ (0.01)$ over all 5 test sections is equal to that in Figure \ref{fig:time} (a), when a previous black section was used to train the best selected model and test the immediate gray section. Thus, the developed model is robust to data variation over time. However, it can be observed a slight decay in F1 score in the last two chunks, suggesting that retraining can be used to maintain performance.

\section{Conclusions} \label{s:conclusions} 
This work focused on evaluating tree boosting methods on balanced and imbalanced datasets of different sizes. Imbalance-XGBoost \citep{wang2020imbalance} was briefly evaluated, and most experiments were performed on a method based on XGBoost \citep{chen2016xgboost}, which stands out in various benchmarks as a recommended boosting system \citep{chen2016xgboost, hajek2022fraud}.

The proposed method scales numerical values between 0 and 1, and encodes categorical data, giving a reserved value when unseen categories appear in test. Preliminary experiments showed that scaling numerical values does not impact small and medium datasets as in theory expected but improves performance when a dataset reaches 100K samples. Therefore, scaling numerical values were used throughout the experiments. Besides, as expected, this report empirically demonstrated that the method increases its detection performance as the dataset size increases. Moreover, the experiments showed that the method's performance decreases as the data becomes more imbalanced, but it is still significantly better than the baseline of precision-recall.

Since the model performs best when dealing with balanced datasets and is affected when data is imbalanced, we tested sampling techniques and classifier optimization to improve detection for imbalanced distributions. In general, sampling to balance the training set deteriorated classifier recognition. Similarly, this result is supported by related studies that found that sampling techniques proved either ineffective or even harmful, deteriorating classifier performance \citep{kim2022, hajek2022fraud}. Classifier optimization, in turn, improves recognition for imbalanced data distributions. We found that Imbalance-XGBoost \citep{wang2020imbalance} can be used to overcome the curse of classification for imbalanced data distributions if the dataset does not have missing values, but in our setup, it was incompatible with Scikit-learn pipelines \citep{pedregosa2011scikit}, and therefore, we did not use it further. In our experiments, the largest improvement was seen with the optimization of XGBoost's $scale\_pos\_weight$ parameter for small and medium-sized datasets, and in the most imbalanced distributions of 75\%-25\% and 95\%-5\%. 

Given that expert inspectors noticed that new anomalous patterns emerge occasionally, we tested the method's robustness over time. The method is robust over time up to some point. When recognition starts deteriorating, retraining is recommended. Although not reported in the experiments presented here, we found that in production, there is a significant difference between the Vanilla version and the optimized version of the method when the data volume is large. Therefore, we recommend caution with hyper-parameter optimization depending on data volume and distribution. 

\section*{Author contributions}
G.V. designed the experiments, wrote the paper, including tables and figures, created the dataset partitions, developed the method, trained, tested it also in production, and analyzed the results. M.W. set the scope, commissioned and supervised the project. A. D. helped bring the model to production and monitored it in production. A.S., S.D., and A.D. provided insights on previous implementations. A.S., K.S., and V.J. collected the dataset for the experiments. 
 
 \section*{Acknowledgment}

We would like to thank Rafael Niegoth for providing business knowledge and assigning the task of developing and evaluating a detection system. We would like to thank Praveen Maurya and Steffen Wenzel for their support with the servers. We would like to thank Lilac Ilan, Krystian Garbaciak, Melanie Mangum, and Bridget Johnson for the feedback on an earlier version of this paper. Finally, we thank the anonymous reviewers for their comments and suggestions.
\printcredits

\bibliographystyle{cas-model2-names}

\bibliography{References.bib}


%
%

\end{document}